# What is Left for Us?

## Second Scholarship Against the Degradation of Research by AI

Claudio Novelli[a], Luciano Floridi[a,b]

[a] *Digital Ethics Center, Yale University, 85 Trumbull Street, New Haven, CT 06511, U.S*

[b] *Department of Legal Studies, University of Bologna, Via Zamboni, 27/29, 40126, Bologna, IT*

**Abstract.** We argue that generative AI can degrade research by eroding the very practices through which scholarly judgement is formed. As constitutive conditions for the production and validation of knowledge, these practices cannot be reduced to the outputs of research, which is what AI tools so effectively simulate. When scholars delegate central tasks of inquiry to Large Language Models (LLMs), they may stop engaging in these practices and lose the formation they provide: the output may improve while the researcher fails to develop. Against this risk, merely keeping humans "in the loop" as prompters or quality-checkers of AI outputs is insufficient to preserve research as a site of intellectual formation. What is needed instead is a renewed commitment to research as a lived practice in which judgement is formed gradually, often through friction, and participation in a scholarly community. We defend it because it rests on four sources and warrants of research that cannot be automated: tacit knowledge, personal commitment, socialisation, and deep reading. These enact what we call 'second scholarship', i.e., the re-appropriation of scholarly craft, chosen out of a critical experience of what AI tools can and cannot do. Researchers who practise it may delegate tasks to these tools, provided that their own judgement in the delegated competence is already formed, that they can answer for the output, and that they keep exercising the competence themselves. What cannot and should not be delegated becomes what research communities must value and answer for. This is what is left for us.

## 1. Introduction

Generative AI (GenAI) typically works in text — most notably in the form of Large Language Models (LLMs hereafter) — and text is the medium through which scholarship is conducted, communicated and assessed. GenAI designates systems trained on very large datasets to generate new content (text, images, code) that is statistically similar to the data on which they were trained. LLMs are its text-based subset: neural networks, today mostly based on the transformer architecture (Vaswani et al. 2017), trained to predict the next token (roughly, a word or part of a word) in a sequence, and then fine-tuned, often with human feedback, to produce outputs that people rate as helpful (Ouyang et al. 2022).

Earlier cognitive artifacts left that medium largely untouched. Libraries and search engines accelerated tasks lying alongside or downstream of intellectual work (e.g., finding, retrieving, sorting). GenAI tools, and in particular LLMs, do these things too, but they also reach into the work itself, performing acts that were once the heart of inquiry, such as framing an argument and reading another scholar's text.

In this article, we argue that this reach carries a cost that no improvement or optimisation in output can offset. More specifically, our central claim is that even when GenAI assistance improves individual research outputs,[1] it may still degrade the formative conditions of research. In fact, when researchers delegate central acts of inquiry to LLMs, they are no longer doing the work through which scholarly judgement is formed. The individual result may look the same, or even better, but this has limited value as research is never only the production of deliverables. What is instead a foundational component of research is the formation and exercise of judgement. It is the slow constitution of a person who can ask the right questions and stand behind the answers, closer to what the Bildung tradition meant by the 'cultivation of a mind' (Sorkin 1983), and to what Aristotle called *phronesis* (or practical wisdom) (Aristotle 2009, 1140a24–b30), than to any particular competence.

Whereas a skill can be acquired and transferred, formation is cumulative, personal, not separable from the person who undergoes it, and thus becoming expertise. A philosophical argument, a legal interpretation, a historical reconstruction, an experimental design and even a logical theorem or a mathematical proof become knowledge only within practices of assessment (and answerability).[2] Even in fields where data, experiments or formal methods dominate, their actual meaning depends on the researcher's trained judgement: e.g., knowing which questions are worth asking, which results matter, which anomalies should trouble us, and which explanations deserve belief.

This is precisely what may fail to develop when GenAI is used for research: the person who would have produced it. So even when the scientific outcomes do not decline, what it is to be a researcher changes, sometimes for the worse. At the scale of a discipline, the danger is sharper since a field that bypasses formation will lack the very judgement on which any check of the technology depends.

Against this risk, we defend the lived, situated and temporally extended experience of becoming and being a researcher. This can be seen as a form of re-embodiment of research. To automate and delegate that experience away is to mistake the research output for the inquiry that produces it, and it has nothing to do with efficiency.[3] In fact, as we will argue, re-embodiment rests on four sources and warrants of research that resist full automation or delegation: tacit knowledge, personal commitment, socialisation and deep reading. We also identify a specific risk we call *hermeneutic drift*: the progressive displacement of a source text's meaning across iterative exchanges with an LLM tool.

Here, our answer to the question in the title begins to take shape. Let us call the pre-critical practice of research, as it was cultivated before AI, *first scholarship*[4]: the patient, quasi-artisanal work of reading, judging, experimenting, calculating and writing, refined without suspicion of its own foundations. GenAI, in general, and LLMs, in particular, threaten the standing of that practice, mainly because a model can return the output while the formative work leading to it comes to seem dispensable.

Faced with this, a researcher can stand in one of three places. Before AI, first scholarship was simply lived, in innocence. Against AI, it is defended but in a posture the threat itself dictates. After AI, the researcher who has used these tools and learned what they can and cannot do returns to the craft deliberately, going through the threat rather than around it. This third state, following Ricoeur's *second naïveté*,[5] is post-critical, and it is what we name *second scholarship*. Re-embodiment is its enactment, the practice that constitutes it. We have inherited first scholarship and we must now develop second scholarship and its methods.

For researchers formed before AI, this sequence is biographical; for those entering the field now, first scholarship may no longer be available as innocence, and its formative content must be deliberately protected in training (see §6.1) if second scholarship is to have something to re-appropriate. What is regained is an engagement with the work and its outcome for which the researcher is fully answerable.

Economists of technological change have long observed that automation tends to raise the value of the tasks it cannot perform, and that complement it, typically tasks that draw on tacit knowledge and judgment and that fall under what Autor (2014), with reference to Polanyi, calls 'Polanyi's paradox' (see also Autor 2015). Nothing guarantees, however, that institutions will recognise this shift, or that its benefits will be evenly shared (Autor 2015). To avoid this, and if what cannot be automated in research is the formation and exercise of judgement, then academia's incentive structures will have to adapt, as today they mostly reward output but may need to reward what is harder to automate, and to do so without turning it into a new privilege of the already advantaged (§6.2).

## 2. GenAI and the Optimisation Ideal

The evidence for the widespread integration of GenAI and LLMs into scientific practice is now substantial (Liang et al. 2025), with early evidence of growth in the production of scientific papers, at least at the preprint stage (Kusumegi et al. 2025). In Wiley's ExplanAItions report, the share of researchers who said they use AI in some aspect of their work rose from 57% in 2024 to 84% in 2025; 62% said they use AI specifically for research and/or publication tasks (Wiley 2025).

These developments have raised serious concerns about their application to empirical research. For instance, the combination of LLMs with large, publicly available datasets – notably biomedical ones – has been associated with a proliferation of low-quality or formulaic research output (Naddaf 2025). Similarly, the use of LLMs to generate synthetic data as a substitute for primary quantitative survey methods has been criticised: both Bisbee et al. (2024) and Westwood (2025) document substantial divergences between LLM-generated responses and actual human survey data, with Westwood further identifying the capacity for synthetic respondents to actively distort polling outcomes (Bisbee et al. 2024; Westwood 2025).

One might wonder why the use of LLMs in research should be considered a problem at all. The case for complacency is not without surface plausibility since scientific investigation is, arguably, fully or at least substantially automatable: delegable to an LLM, with the scholar's role acting as a prompt generator or, in the more radical version of the argument, dispensed with entirely. Why not, then, hand the enterprise over? One could even imagine a complex multi-agent system made of models trained on scientific texts and assigned synthetic deliberative personas (Chapuis et al. 2022; Yang et al. 2025), capable ofprocessing abstract and concrete research questions alike, generating solutions and objections in indefinite succession. The perfectly optimised symposium, running without interruption or fatigue.[6]

Yet, we maintain that the wholesale replacement justified by the optimisation ideal of automation would be undesirable for two principal reasons. The first concerns the reliability of scholarly outputs at the level of individual claims: human involvement – what the AI governance literature terms keeping "humans in the loop" (Lazaros et al. 2026) – continues to play a demonstrable role in ensuring their accuracy. We address this concern in the following section (§3). The second operates at a more structural level: research needs to preserve its formative and communal practices through which scholarly judgement is developed, claims are tested and trust is built. The second reason is the subject of §4.

In the vocabulary of the Introduction, the optimisation ideal is what ends the innocence of first scholarship since once a deliverable can be had without the formative detour, the detour must be chosen instead of being inherited unreflectively. Second scholarship names that choice.

## 3. Inside the Loop: Is That Enough?

A growing body of scholarship argues that AI-generated scientific output is not reliable without some form of human input, revision or verification. Agrawal, McHale, and Oettl (2026, NBER WP 34953, pre-publication at the time of writing), for instance, offer a useful framework by dividing scientific production into four main phases (i.e., question generation, idea generation, design generation and testing) and showing both where AI performs well on its own and where human judgement remains necessary to ensure reliable results. They define judgement as the capacity to discern and evaluate the practical and normative significance of actual and possible states of the world, including sensitivity to context, the ability to weigh conflicting objectives, causal hypothesising and the capacity to draw analogies (Agrawal et al. 2026, 4–5).

In general, they argue that human judgement remains indispensable across all four stages of the scientific workflow, though its necessity is most acute in data-sparse environments and conceptually demanding tasks, conditions that arise wherever theory outruns measurement, evidence is scarce or framing assumptions are themselves contested (e.g., in humanities, law or philosophy).

Their analysis becomes more fine-grained when they turn to the individual stages. In question generation, it is true that AI has made hypothesis generation remarkably easy, but a strong research question still implies that there is something to be surprised by (Agrawal et al. 2026, 10).[7] Whether a researcher is surprised, however, depends on what the authors call a “prepared mind”, e.g., one already familiar with the state of the art; and when an algorithm produces a long list of anomalies or patterns, human judgement is needed to distinguish what is genuinely significant from what is merely noise (Loru et al. 2025).

In idea generation, the authors lean on Polanyi's concept of tacit knowledge (Polanyi 2009) to suggest that humans excel at the kind of abductive inference that AI currently struggles to replicate, not least because the creative “guess” involved in formulating an explanatory hypothesis also requires a degree of personal commitment from the scientist (Agrawal et al. 2026, 18–19). Although LLMs can generate a vast range of recombinations of existing ideas, human judgement remains essential in navigating the combinatorial search for genuinely meaningful explanations.

Even in design generation, where AI's promise is most evident (e.g., AlphaFold), given that a design may need to satisfy multiple objectives, making choices amid trade-offs can require scientists' judgement and personal commitment.

Finally, in the testing phase, given the often black-box nature of highly flexible machine-learning-based function approximators, scientists' judgement might become even more important for validating that results represent genuine discoveries rather than computational artifacts (Agrawal et al. 2026, 24–25). But this is not surprising, as we anticipated in the introduction.

To these reasons for the indispensability of human input, we should add another, of a different kind: the maintenance of a "progressive" research programme. Most AI systems, and LLMs in particular, are structurally oriented toward pattern-completion within existing data, producing a gravitational pull toward "conservative" outputs, what the philosophy of science, following Kuhn, might recognise as a bias toward *normal science* (Kuhn 1962) at the expense of paradigm-breaking inquiry.

The mathematician Terence Tao, in a recent public conversation on AI and scientific discovery (Tao 2026), used the history of astronomy to illuminate a related risk. He noted that Ptolemy's geocentric model had been refined over a millennium through increasingly elaborate modifications, becoming remarkably precise (and yet was fundamentally wrong). For this reason, when Copernicus proposed the heliocentric alternative, it was *initially* less accurate than the Ptolemaic system it sought to replace. As Tao observed, this happened because an incorrect but fully developed and refined theory can appear more compelling than one that is correct but only partially developed.

This risk is not unique to AI, but AI may amplify it by intensifying existing optimisation pressures within scientific practice. Human scientific institutions, such as peer review and funding structures, arguably already favour what Lakatos called *degenerative research programmes*: accumulating protective modifications around a core hypothesis or theory, making it simultaneously incapable of generating new predictions and increasingly difficult to question (Lakatos 1970). Likewise, LLMs trained on the outputs of these same institutions (and practices) inherit their blind spots and, by optimising for what is already well-represented in the data, may systematically underweight those anomalies and outliers from which paradigm shifts historically emerge. The space for disruptive ideas does not disappear, but it narrows, especially in an environment flooded with AI-generated, formally accurate, machine-generated hypotheses. Even if this conservatism concerns the direction of a research programme rather than the development of a researcher, it bites hardest where the trained judgement that would recognise a genuine anomaly has itself been allowed to atrophy (which is the formation worry we develop in §4).

A related and, we think, deeper point is that the degenerative research dynamic is not imposed on academic practice from outside, from some economic power, or from civil society. It is generated by the social organisation of academic research itself, for instance, its reward system. What AI changes is not the direction of these incentives but the cost of compliance with them. LLMs collapse the marginal labour cost of producing a formally acceptable paper, leaving the incentive gradient intact while removing nearly every friction that once limited how far individuals could travel along it. For this reason, it is essential to address institutional aspects, as we also suggest in §6.

## 4. Beyond the Loop: Re-embodying Research

That LLMs make writing easier is by now a commonplace, and it has made us worry that writing, as a mode of thinking, is under pressure. A quieter crisis concerns reading. An undergraduate who wants to grasp Wittgenstein's picture theory of language no longer needs to read the Tractatus; they can ask an LLM for a summary, or upload the text and ask for its key claims. That is faster. It is also a different act of the mind. Nor is the summary a neutral stand-in for the text: when 4,900 summaries of scientific texts produced by ten widely used LLMs were compared with the originals, most models generalised the findings more broadly than the sources warranted, even when explicitly prompted for accuracy (Peters and Chin-Yee 2025). What both shortcuts cost is a kind of epistemically productive friction (Sher 2010): the work of following an argument closely, drafting and redrafting a claim, answering an objection. Such friction matters because it is one of the practices through which judgement is formed and claims are tested.[8]

The consequences extend in two directions. The first is how authors engage with the literature: whether they read sources closely or merely process them (for related correlational evidence on cognitive offloading, see Gerlich 2025). Of course, scholars have always read selectively and synthetically; this is not new. But AI accelerates that tendency to the point where the original source may not be consulted at all. The second direction is peer review. If reviewers lean on LLMs, or read less carefully, the community's main check on accuracy weakens at its core (Zhu et al. 2025; Mann et al. 2025).

Set beside the previous section, the picture is troubling. Researchers risk becoming mere contextualisation machines or prompt designers. Push the "optimisation" of reading and writing far enough, and the work begins to detach from the critical thinking and the intellectual biography that have always formed scholars. Worse, that optimisation erodes the conditions under which a researcher's future self is formed, i.e., bypasses the lived experience of inquiry, and there is less and less on which to build a scholarly self. The capacity to take a position and defend it grows through the friction of genuine encounter with ideas. A workflow that strips that friction away does not leave the formation intact while making it faster. It replaces it with something else.

The question is not whether efficiency has value. It is what efficiency leaves for us, and whether the lived, formative experience of inquiry is still worth defending. We believe that there remains not only room, but an increasing need, to defend the *re-embodiment* of research, that is, a lived, personal, and communal practice, made of genuine encounters with texts and colleagues and of the slow, frictional formation of an intellectual commitment that is existential as much as methodological. Living the work, rather than delegating it, enriches (the experience of) research. More to the point – and here we mean to resist paternalism – the quality, depth and originality of the work itself depend on the very conditions of formation that AI's degenerative optimisation dismantles. The next subsections offer four arguments in support of re-embodiment and, by doing so, specify what second scholarship must preserve.

Re-embodiment is not a call to go entirely offline or unplug. As a prescription, unplugging fails on three counts. It will not happen, and a programme that waits for it is idle. It is beside the point: the harm is not that researchers use LLMs, but that they let the technology do their formative work. And it is confused: in an onlife world, there is no offline to return to. The line between online and offline has dissolved, and we are, in the relevant sense, informationally embodied organisms inhabiting a single, hybrid environment (Floridi 2014; Floridi 2015). Second scholarship is for this reason the onlife form of the craft as it does not aim to return to an offline that no longer exists, but reclaims the formative work within the hybrid environment.

The case for re-embodiment finds a foundational ally in Dewey, who stresses how much inquiry is initiated and sustained by felt perplexity (Dewey 1910, 201), which is, from our perspective, the embodied sense of being thwarted that directs a reasoner's attention to a problem in the first place, defines it, and lets one know, at the end, when it is resolved (Douglas 2024, 21). Without that felt sense, there is no problem detection, only error detection. Re-embodying research, in this light, is much more (and less) than a sentimental gesture toward the lived experience of scholarship but a structural requirement of the activity itself. The four arguments that follow are, in different ways, specifications of that requirement.

### 4.1. Tacit Knowledge

We have seen (§3) how tacit knowledge is often invoked in defence of the irreplaceability of the human researcher. As Polanyi famously put it, "we can know more than we can tell" (Polanyi 2009, 4). This is a structural thesis about how knowledge, including scientific knowledge, works. Whenever we attend to something, we rely on background elements that support our judgement without themselves becoming the object of attention (Polanyi 2009; Turner 2022). Polanyi calls the object of attention the focal object and the supporting background the subsidiaries. We know, in other words, from the subsidiary toward the focal (Polanyi 1968, 29).[9]

At the same time, the tacit-knowledge argument needs qualification. In its strongest sense, tacit knowledge is easiest to see where competent performance depends on bodily ways of doing things: the classic TEA laser case, in which experimental know-how was transferred only through close practical contact with experts who themselves could not say what they were doing differently to make it work, remains paradigmatic (Collins 1974).

Yet, much of what is called expert judgement in the humanities is not tacit in this strong sense. Its basis is openly discussed (e.g., in supervisions and disputes), and the difficulty is more about richness, meaning that there is always more in the formed scholar than any single articulation can convey.

For this reason, we think that the humanities have their own analogue of tacit knowledge, a form adjusted to a discipline whose medium is interpretation rather than physical manipulation. We may call it an *epistemic disturbance*: the felt sense that something is missing or wrong, that a view has not captured something important, that one disagrees and needs to investigate further to find out why.[10] Drawing again on Dewey, this felt sense is what initiates and sustains inquiry. Without an embodied capacity to register that something is amiss, there is nothing to direct attention to a problem and nothing to keep one with it. We believe, though, that epistemic disturbance, also for this directional focus, still counts as tacit knowledge in a Polanyian sense, i.e., it is known from the subsidiary toward the focal, operates before it is thematised, and cannot be fully made explicit.

In both senses, however, the tacit dimension consists of cues, habits, embodied sensitivities, and practical judgements that underwrite explicit thought without ever becoming fully thematised. Explicit knowledge always depends on these internalised capacities that exceed what can be articulated. This matters especially in the early, generative stages of scientific inquiry, where judgement and skill often operate before formalisation.

Tacit knowledge resists codification. What functions as a subsidiary clue cannot simply be brought into focal awareness without changing its role. Once attended to directly, such knowledge ceases to operate as part of the background competence that made action or judgement possible in the first place (Autor 2014). In practice, this is why tacit knowledge is often transmitted person-to-person through experience, imitation, and implicit learning (Howells 1996).

A deeper question is whether its resistance to codification is merely practical or principled. The stronger claim is that it is principled. Any explicit rule requires interpretation in order to be applied, but interpretation itself depends on tacit knowledge. If one tried to make that tacit knowledge explicit as well, it too would require interpretation, and so on, without end. This regress argument, indebted to Wittgenstein's reflections on rule-following, suggests that, however much is written down, tacit knowledge is merely displaced one step further back (Thornton 2006, 7–8).

A distinction between types of tacit knowledge helps clarify easy and strong cases of tacit knowledge resistance to codification (Collins 2013). *Relational tacit knowledge* is the least difficult case: conversational norms, social expectations or professional etiquette may remain tacit because of convention, secrecy or lack of need, but they could in principle be made explicit (Collins 2013, 11). *Somatic tacit knowledge*, grounded in bodily capacities – e.g., balance, muscle memory, the practiced hand – could in principle be described by science, but description is not transmission (Collins 2013, 31). The most recalcitrant case is collective *tacit knowledge*: knowledge carried by immersion in a community over time, by the accumulated sensitivity of someone who has simply been around long enough (Collins 2013, 133). Fully describing it would require a vantage point outside the very practices being described. And no such point exists.

The point is that scientific judgement depends heavily on this collective dimension. A trained scholar can identify a fringe article almost instantly, not by consulting explicit criteria but through a socially acquired sense formed through apprenticeship, conversation and prolonged disciplinary immersion.

This is precisely what LLMs cannot extract from text alone. No matter how vast the corpus, it captures only what has been externalised: the published outcome of a process whose formative stages often unfold earlier, in unrecorded conversations, informal exchanges, and long-standing relations of trust and authority (Collins and Thorne 2026, 6).

Giving LLMs access to recordings, or even trying to turn parts of tacit research knowledge into machine-readable material (Liu et al. 2026), does not solve the problem because the issue here is not that the right words are unavailable. It is that words mean different things depending on who says them, to whom, at what stage of a relationship, within what community, after what shared history. Scientists are not exposed to discourse as random fragments. They go through a structured initiation, in which the knowledge accumulates in a specific order, alongside people who implicitly sort the important from the trivial: e.g., undergraduate training, a PhD under a specific supervisor, years of working alongside people who implicitly sort the important from the trivial, the live debate from the settled question. What an LLM would extract from recordings is, instead, fragments arriving without the same sequence of acts that builds up a (good) scientific discourse, and without the social relationships that give them weight (Collins and Thorne 2026, 15). An LLM can, at the state of the art, produce a linguistic appearance of competence through surface pattern-matching but without the underlying tacit substrate that would make such competence integral to practice. Call this, if one needs a label, *pseudo-interactional expertise*: the linguistic surface of interactional competence without the socialised substrate that would normally support it.

### 4.2. It's You and Your Research! On Personal Commitment

A second reason to resist reducing researchers to components in an AI-optimised production loop is that research is not merely a procedure for generating outputs. It is also a process through which a person becomes a certain kind of knower. Someone with a specific intellectual identity who brings formed capacity to the society in which they live.[11]

Polanyi's account of scientific knowledge insists that genuine inquiry requires personal commitment, which is deeply connected to tacit knowledge. This commitment forms part of what Polanyi calls his "fiduciary programme": the overarching epistemological claim that all knowledge is ultimately grounded in an act of belief that cannot itself be fully justified by explicit reasoning (Polanyi 1958, 264–268). To commit to a claim in this sense is to engage one's full person – the rational faculties but also one's intellectual passions, heuristic drives and aesthetic sensibilities – in an affirmation that is genuinely at risk. This makes the claim genuinely answerable, that is, staked by someone whose credibility and judgement are implicated in its standing.

Without personal commitment, it becomes harder to stand behind an argument as one's own, and when scholars can no longer do that convincingly, their most important asset, i.e., reputation, is eroded. The authority of a legal scholar's interpretation, a philosopher's argument, or a historian's reconstruction rests heavily on the community's confidence that the person behind it has genuinely reasoned their way to it. And, accordingly, they can be questioned, challenged, and, if necessary, held accountable.

Personal commitment is the standing posture of a researcher who has felt and inhabited enough of what §4.1 called epistemic disturbance to take a stake in resolving it. The two are not separable in practice: the felt sense that a view has not captured something important is what gives one something to be committed to, and committing to a claim in turn exposes the scholar to further occasions of being thwarted by it. To outsource the affective layer of inquiry is, accordingly, to weaken the conditions under which commitment can form at all.

In fact, personal commitment is rarely frictionless. It is often forged through the difficulty of making a point one can confidently own. And this is where a point about epistemically productive cognitive friction - something that risks being lost in AI optimisation pressure - belongs. Difficulty is not intrinsically admirable, nor is every obstacle in research. Some forms of difficulty, however, improve learning and thus contribute to the formation of the researcher. Work on "desirable difficulties" shows that conditions which slow apparent performance can nonetheless produce qualitatively superior - i.e., better retained information in the long term and more robust judgement - and more transferable understanding (Bjork and Bjork 2011; Bjork and Bjork 2020). The difficulty of constructing a philosophical argument from scratch, of reading a resistant legal text until it yields its structure, of defending an interpretive position under sustained attack from interlocutors who will not be easily persuaded, should not be seen as inefficiencies. They are part of the process through which the scholar develops the kind of epistemic judgement that academic inquiry requires. A judgement that is no less operative, if often less conspicuously, in the design and assessment of lab experiments or the interpretation of data. Outsourcing these activities to an LLM does not produce the same scholar faster.

The combination of these aspects unveils what is, frequently, a substantial part of the social value of a researcher. Much scholarly work does not have a quantifiable social impact and, even less, one that can be measured independently of the process by which it was produced. What matters is the presence in society of people who have spent years forming an educated, committed, defensible engagement with specific questions. The scholar is someone whose sustained encounter with those subjects has formed a particular kind of mind: capable of discrimination and accustomed to holding complexity. The value of their work lies partly in the kind of person they have become through research, and in the fact that this person can now intervene in teaching, public debate, institutional life and cultural self-understanding with more than just amateur enthusiasm. Optimising research through AI risks producing fewer people of this kind.

Early empirical signals are consistent with this concern, though they should be read with care. A much-discussed but non-peer-reviewed preprint (Kosmyna et al. 2025) reports EEG evidence of reduced brain connectivity among essay writers using ChatGPT and a diminished sense of ownership over their texts. This preprint should be treated as provisional: the sample is small (n = 54 in the main sessions; n = 18 in the critical fourth session), institutionally homogeneous (Boston-area university students, ages 18–39), and the task (SAT-style essay writing) may not generalise to research writing. The direction of the findings is suggestive and consistent with the "desirable difficulties" framework, but stronger peer-reviewed evidence will be needed to settle the question.[12]

### 4.3. And Yet, It's Not Just Your Research: Socialisation

Tacit knowledge and personal commitment operate at the individual/cognitive level. Complementary to these arguments is a distinct claim that operates at the community level. Scientific knowledge emerges from socialisation, that is, structured and constant community practices, rather than from just aggregating individual outputs. Expertise is transmitted this way as well (Lave and Wenger 1991). The conditions under which inquiry yields something robust and defensible are themselves irreducibly social. Examples are public venues for criticism, shared evaluative standards, and relative equality of intellectual authority among participants.

One might suppose that science achieves robustness precisely by keeping its internal epistemic norms (or *constitutive* values) – e.g., accuracy, simplicity, predictability – insulated from the social world and thus from factors like political commitments or cultural assumptions. But this insulation is quite controversial (Longino 1990). Because any hypothesis requires background assumptions (or *contextual* values) to connect it to evidence, and those assumptions are not themselves directly testable. Values from the broader social environment enter scientific reasoning invisibly, without violating any explicit methodological rule (Longino 1990, 51).[13]

If contextual values enter scientific reasoning invisibly through the social environment, they can, over time, crystallise into the institutional machinery by which research is evaluated and rewarded. In fact, the same social dimension of research and the social pressure or competition that comes with it, which gives research its epistemic robustness (e.g., peer venues and shared evaluative standards), has also produced the metric architecture of contemporary academia, such as the 'publish or perish' culture.

The social character of knowledge production becomes especially salient in collaborative research, and in contemporary science, collaboration is less an option than a contingent necessity. When this is the case, socialisation manifests as an epistemic dependence relation between researchers. Empirical work on collective intelligence and diversity suggests that socially diverse groups, especially when they engage in iterative and responsive forms of inquiry, can outperform isolated individuals or more homogeneous groups in solving complex problems (Page et al. 2019). This is also explained by the fact that collaboration can help reduce the impact of individual biases. People are often much better at noticing weaknesses, fallacies, and blind spots in someone else's reasoning than in their own (Grossmann 2021, 5; Mercier and Sperber 2017). In a way, therefore, science exploits and "reorganises" a basic feature of human cognition, since we evolved to learn with others and often revise our beliefs through social interaction (McIntyre 2019; Grossmann 2021, 5).

A broad consensus in social epistemology holds that this relation should be more than reliance; it should be genuine trust (Hardwig 1991; Goldberg 2011; Rolin 2020). And trust, unlike mere reliance, can be betrayed. It therefore requires, as its necessary condition, that parties are moral agents, accountable for their epistemic conduct, especially within their scientific communities (Goldberg 2020). This holds for internal collaboration within research teams and for external collaboration: that form of dependence that occurs whenever a researcher builds on the published work of others whose methods and integrity they cannot independently verify.

LLMs lack the conditions for genuine participation in these social practices on both grounds (robustness and collaboration) simultaneously. A first obstacle to socialisation concerns how LLMs encounter knowledge. In research, we have seen that constitutive values are themselves products of socialisation (and contextual values). More specifically, they are strongly influenced by that kind of socialisation that happens through immersion in small, bounded groups that rely on face-to-face communication, like the family (the so-called primary socialisation) (Collins 2025, 1251). This matters for research science because, as also indicated by developmental evidence, this kind of face-to-face, caregiver-mediated interaction is the primary mechanism through which children acquire generalisable cultural knowledge and calibrate the basic disposition to treat communicated content as reliable or unreliable (Fonagy and Allison 2014; Csibra and Gergely 2009). It is through this same relational context, and not formal instruction, that humans develop the foundational orientation toward what counts as trustworthy communication, and are socialised into the norms of honest conduct that underpin it (Talwar et al. 2022; see also, on the shared-intentionality basis of these capacities, Tomasello 2018).

By contrast, an LLM trained on internet text bypasses this formation entirely, inheriting the outputs of socialised humans without undergoing the formative process that produced them. Attempts to compensate through reinforcement learning, with human labelling and feedback, resemble retrospective socialisation: developers acting as surrogate parents, installing a moral compass after the fact in a system that has never undergone the formation of such a compass, presupposes (Collins 2025, 1257–1258).

But AI also fails to meet the accountability condition structurally. LLMs cannot be held morally responsible and cannot suffer the genuine consequences or sanctions that give accountability its epistemic function (Resnik and Hosseini 2025). An AI model that participates in the production of scientific claims builds an opaque epistemic dependence, just as with human collaboration, but without being answerable for them, and is not a member of the community whose trust relations make those claims knowledge. AI systems' epistemic opacity is essential rather than accidental: unlike a human collaborator whose methods may be opaque to their colleagues but remain in principle transparent to someone, an LLM's computational processes are opaque even to its developers (Koskinen 2024, 464–465).

It is essential to stress that the social character of research also explains why individual AI harm becomes collective. If the formation arguments of Sections 3.1 and 3.2 hold, then each researcher who delegates the formative work develops a thinner judgement and because the community's capacity to check, referee and correct is nothing more than the aggregate of those individually formed judgements, a discipline that bypasses formation at scale erodes the very capacity on which the reliability of AI-assisted research depends. Socialisation thus shows that second scholarship cannot be a private virtue. Its re-appropriation must be sustained by a community that still contains formed judgement, and it is a collective achievement or it is nothing, which is why §6 treats it as a matter of institutional design.

4.4. Deep Reading (and Writing)

A fourth reason to defend the re-embodiment of research concerns what is sometimes called, in the neuroscience literature, deep reading (Wolf 2018; Wolf and Barzillai 2009),[14] which has key consequences for writing as well. Essential to understanding this notion is the observation that the human brain did not evolve to read (Wolf 2010). Reading is a cultural invention, approximately six thousand years old, that repurposes neural circuits evolved for other functions (e.g., vision, language, motor planning, memory) into what Maryanne Wolf calls the "deep reading circuit" (Wolf 2018). Wolf's research depicts how the deep reading brain enables the development of some of our most important intellectual and affective processes: internalised knowledge, analogical reasoning and inference; perspective-taking and empathy; critical analysis and the generation of insight.

The deep reading circuit is no less operative in scientific inquiry than in literary reading. To engage with a scholarly text properly is to follow an argument across its internal structure, hold its parts together in memory, test local claims against the whole, and gradually build an interpretation that can be owned and defended. Cognitive-neuroscientific work on expository and scientific reading supports this point by showing that successful comprehension depends on such integrative activity. Better readers decode sentences more efficiently, of course, but that's possible as they recruit the cognitive resources necessary to construct a coherent mental model of the text, also to maintain relevant information in working memory, and integrate what is read with prior knowledge and memory (Keller et al. 2024; van den Broek 2010).

However, the deep reading circuit is not fixed. It develops with use and weakens with disuse. This means that, when the reading brain skims, as is increasingly common with digital reading, it reduces the time allocated to deep reading processes, and meta-analytic evidence confirms that screen-based reading is consistently associated with lower comprehension outcomes than print (Delgado et al. 2018).

AI-assisted reading is yet another form of reading, distinct from surface-level digital reading. Reading a text by uploading it to an LLM and querying it in natural language is, in one sense, highly interactive: the researcher can ask questions, request summaries, prompt for counterarguments and examples. This is much more targeted reading. But it raises at least two problems: (a) *hermeneutic drift*; (b) interaction does not stand by itself.

On the first problem, by hermeneutic drift, we mean the progressive displacement of a source text's meaning that occurs when a researcher reads through an LLM by posing iterative questions. Since each question by the reader carries an implicit interpretive frame, and given the deferential design features of LLMs (also called sycophancy in literature) (Sharma et al. 2025), that frame tends to be reproduced in the generated response, and each response then conditions the context for the next question. Over successive turns, the account the researcher receives increasingly reflects their own prior understanding back to them rather than the original content of the text itself. According to Buzási (2022), Ricoeur's hermeneutics gives this a precise structure. Understanding a text has two moments: *distanciation*, in which the reader works on the text through a mediating apparatus rather than directly, and *appropriation*, in which the reader takes the resulting interpretation back as their own and can restate and defend it without the apparatus. Interpretation is complete only when the first is followed by the second, since distanciation is a legitimate stage but a defective end. Hermeneutic drift is distanciation without appropriation: the reader engages the text only through the model and never reconstructs the interpretation independently, so what is understood is the model's account, rather than an interpretation that the reader could produce and defend. Second scholarship satisfies the condition that hermeneutic drift fails: the reader may work through the model, but then reconstructs the reading independently and is answerable for it. Let us be clear that we have not measured hermeneutic drift. We develop it based on empirical literature about cognitive bias (and degradation) and sycophancy as mentioned already, but it will need to be accurately tested.

The second problem is that interactivity, by itself, does not make reading deeper. This is instructive to recall in the context of an older debate. Hypertext was often celebrated as a more active mode of reading, yet research showed that the added demands of navigation, choice and visual processing could impair comprehension by overloading working memory and interrupting the construction of a stable situation model. These findings suggest a difference between activity and depth. A reading practice may be interactive and yet remain cognitively thin, either because attention is fragmented across too many stimuli or because the effort required for conceptual integration is removed (DeStefano and LeFevre 2007).

AI-mediated reading produces a formally similar but mechanistically opposite failure, albeit that opposition is instructive. Where hypertext impairs depth by demanding too much of working memory, AI-mediated reading impairs it by demanding too little. The navigational overload of hypertext is replaced by frictionless delivery: the model absorbs the text, distils what the researcher appears to want from it, and returns a fluent response. Readers may be exposed to interactional stimuli, such as asking questions and immediately using a specific piece of information to support their point, without having done the deep reading conceptual work. Again, early empirical evidence is consistent with this picture. Students using LLMs for information gathering reported significantly lower cognitive load than those using conventional search, but this reduction came at the cost of weaker reasoning and argumentation in their conclusions (Stadler et al. 2024; Kosmyna et al. 2025).

## 5. Some Objections to our Argument

Before moving to the constructive, so-what?, part of the paper, let us address two potential objections against our analysis.

A first objection, both against our analysis so far and what comes next, may arrive from that strand in the philosophy of cognitive science which argues that LLMs are best understood as the latest instance of a long sequence of cognitive artifacts (e.g., writing, libraries, search engines, online encyclopaedias) that each, at first, appeared to threaten cognitive capacities and then, in time, were absorbed as extensions of them (Heersmink et al. 2024; Cassinadri 2024). We should be careful about what we are and are not conceding. We do not dispute that LLMs can function as cognitive extensions. They can clearly do so, just as the calculator and the search engine do. Nor do we dispute that, so used, they can also raise the quality of individual scholarly outputs. Our argument is that extension alone is not enough. A science that is cognitively extended but no longer constituted by the lived formation, the socialised judgement, and the communal practices (as we elaborate in §4) is not the science that the extension was supposed to be helping.

The objection has a venerable form, the complaint Socrates raises in Plato's Phaedrus against writing itself, and that complaint is now usually cited as a cautionary tale about misplaced alarm. We accept the caution but deny that it reaches the present case, and a narrower point shows why. Earlier cognitive artifacts mainly supported memory, calculation, and search. They did not, or did so only marginally, generate the very interpretive moves central to academic inquiry. A library does not propose readings of Wittgenstein, and copying Kripke's works on Wittgenstein, which you can find on Google, strongly exposes you to plagiarism allegations. LLMs, by contrast, recombine and reorder existing readings into new-looking wholes, which also makes such borrowing much harder to detect. But the point is even more general: where earlier tools displaced auxiliary cognitive labour, LLMs increasingly simulate what had been the criterial activity of inquiry, whether that activity is the close reading of a philosophical text or the verbal articulation of a mechanism in a scientific paper.

There is one earlier exception, and it is instructive, because the field has been here before. From the mid-1970s, the dominant programme in applied AI set out to build expert systems by eliciting what physicians, geologists and chemists knew and recasting it as explicit rules. It reached directly for the criterial activity, the expert's judgement itself, and it foundered on what Feigenbaum named the knowledge-acquisition bottleneck, i.e., experts could not say what they knew, and the structured interviews, protocol analyses, and rule bases devised to extract it captured the articulable surface but never the tacit substrate beneath (Felgenbaum 1977). The flagship systems in medicine and chemistry, such as MYCIN and DENDRAL, performed impressively within narrow boundaries and degraded sharply at their edges, where the unstated judgement of the expert would have carried a human through. Dreyfus and Dreyfus had predicted the failure. On their account, expertise is not rule-following but a situational, embodied grasp that the novice's explicit rules merely approximate, so that asking experts to state their rules forces them back to the novice's level and yields a rationalisation rather than the competence (Dreyfus et al. 1986). Collins made the parallel sociological case in 'Artificial Experts': the knowledge carried by a collectivity cannot be acquired by a machine that stands outside its form of life (Collins 1990).

The expert-systems programme failed at exactly the point where the criterial activity met the tacit substrate, and the failure has since been inverted rather than overcome. Where the older programme could not get tacit knowledge into the machine by asking experts to articulate it, the training of a language model encodes a statistical residue of that knowledge from the textual traces experts leave behind, so that the model now, in Polanyi's own formula, 'knows' more than it can tell (Kambhampati 2021). This is why the LLM can produce the linguistic surface of competence that the rule-based system never could. But it is also why the worry of this paper is not dissolved by that success. What the model encodes is the externalised residue of formation, not the formation itself; and our claim, developed below, is that the residue cannot substitute for the process that produced it, either in the individual who would have undergone that process or in the community that depends on people who have.

One objection, finally, presses harder than the others, and it is the one on which the formation argument stands or falls. Why should delegation prevent formation rather than merely relocate it? A scholar who critically supervises an LLM's output, the objection runs, exercises judgement in the supervising, much as one is formed by reading and criticising the arguments of others; as such, delegation need not stop formation. The reply is that supervision and production are not equivalent exercises. There is an asymmetry between recognising a good argument and being able to generate one: evaluation draws on a narrower competence, since one can often tell that a result is sound without being able to produce it, and it is the generative capacity, not the recognitional one, that formation builds. Delegation leaves precisely that capacity unexercised, and the gap is now visible empirically. In a randomised study, students who completed a writing task with ChatGPT produced better essays in the short term, yet their knowledge gain and transfer did not improve relative to controls; the authors name the mechanism metacognitive laziness, the offloading of the self-monitoring and planning through which understanding is built (Fan et al. 2025). A complementary survey of knowledge workers finds that higher confidence in a generative system predicts measurably less critical thinking, as the activity shifts from producing a result to verifying the machine's (Lee et al. 2025). The lesson is that supervision can preserve judgement only where enough formation already exists to make supervision meaningful. One cannot supervise one's way to a competence one never built, a point we develop in §4.1. In our terms, supervision without formation is merely an attempt at second scholarship, which fails because there is no first scholarship on which to build.

This is, in contemporary terms, the irony of automation that Bainbridge identified long ago: automating the easy part leaves the operator both less practiced at the hard part and newly burdened with monitoring it (Bainbridge 1983). Vallor has given the broader pattern a name, moral deskilling, the erosion of practical capacities that only their sustained exercise develops (Vallor 2016). Our thesis is its epistemic counterpart. Vallor's concern is the habituation of virtue; ours is the formation of judgement, which we ground not in habit but in the maker's knowledge that only doing the constitutive work confers.

Let us clarify that scholars who have already built the relevant competence can use these tools to check, compare, or accelerate specific tasks without forgoing second scholarship, because the recognitional capacity they exercise in supervising rests on a (generative) capacity already formed. But two conditions qualify the concession. The first is that the threshold is task-specific, that is, expertise in one area does not license supervision in another, and the formed scholar becomes a novice again at the edge of their field, which is often where research happens. The second is that formation is not a one-off endowment but a capacity maintained through exercise. In a simulator study of airline pilots, manual control skills were largely retained under automation, but the cognitive skills needed for manual flight, such as keeping track of the aircraft's position and recognising instrument failures, showed more frequent and significant problems, and their retention depended on how actively pilots stayed engaged while the automation flew (Casner et al. 2014).

## 6. What Is Left for Us: How to Develop Second Scholarship?

The four arguments for re-embodiment presented above do not become operative on their own. Without structural support, they remain normative aspirations easily overridden by existing incentive pressures. Some concrete measures to contrast the uncontrolled use of AI in research have already been adopted, but we are unpersuaded. Journal requirements that authors disclose how AI was used in producing a paper are a representative case. Such disclosures are cheap to produce, hard to verify, and, more importantly, leave the researcher-in-the-loop paradigm fully intact.

The way forward is neither to entrust research to the technology nor to retreat completely from it. This is a single demand with two faces, and they are precisely the two abstractly named: to disrupt the default to trust, and to build the institutions that keep the scholarship embodied. The two faces track the two movements of the second naïveté from which second scholarship takes its name: critique, the deliberate interruption of innocent trust (§6.1), and re-appropriation, the reasoned return to the practice (§6.2). The first concerns the stance a researcher takes toward the tool in the moment of use; the second rebuilds the embodied and communal conditions of formation that the four arguments defended, and makes them count where careers are decided.

### 6.1. Disrupting the Default to Trust (*Inside the Loop*)

Whoever consults a model begins, by default, in a posture of trust: the fluent answer arrives as something to accept, lightly correct, or at most spot-check, rather than as a claim to interrogate from the position of someone who could have produced it. This posture is what Nguyen calls trust as an unquestioning attitude (Nguyen 2022): relying on a resource while suspending deliberation over its reliability, letting it through the gate that monitoring and challenge would otherwise hold. Extended too readily, the unquestioning attitude becomes what Nguyen calls agential gullibility (Nguyen 2022, 237), i.e., the hasty integration of an external resource into one's own thinking, and that is precisely the formation risk. People end up not weighing each output on its merits and instead settle into a general heuristic of following the machine (Buçinca et al. 2021). The deference that drives hermeneutic drift (§4.4) and the metacognitive laziness identified in §3 both presuppose this default; weaken it, and much of the rest loses its grip.

The remedy follows. If the trouble begins with a default to trust, the way out begins by disrupting it, though not, we should say at once, by installing blanket distrust. A researcher who rejects every output is the mirror image of one who accepts every output, and human-factors research established long ago that the two failures, misuse born of over-trust and disuse born of distrust, call for a single correction, which is calibration rather than suspicion (Lee and See 2004).

There is a deeper reason to disrupt the default, and we laid its foundation in §4.3. Trust in the agential sense, the kind that can be betrayed, is owed only to someone who can be held to account, and a model is not such an agent (Koskinen 2024); to extend it that trust is, strictly, a category mistake. The unquestioning attitude, by contrast, is all too easily extended to a tool, and it is this, not some impossible agential trust, that needs disrupting. What must be withheld from the model is not the trust owed to a colleague, which was never available, but the suspension of monitoring that we grant a reliable instrument and that a fluent model invites us to over-grant. Disrupting it returns the researcher to the one position the machine cannot occupy, that of the agent who is answerable for the claim. This, in a phrase, is what is left for us: not the production of the text, which can be delegated, but the answerability for it, which cannot.

One might think the task easy, given how fragile trust is, slow to build and quick to lose, so that a single exposed error could undo it. We are less sanguine. As the systems grow more fluent, the visible errors that would prompt spontaneous correction grow rarer while the deference they invite grows stronger, so the disruption cannot be left to the accident of a caught mistake. It must be designed and rewarded.

At the level of the individual scholar, what we are after is not independence from the tool but intellectually virtuous autonomy (Carter 2020): a self-direction compatible with reliance and outsourcing, provided the agent retains control over which dependencies to enter and when to reopen them. Calibrated reliance is that autonomy made specific to the model. Disrupting the default cannot be a matter of exhortation, since telling overworked researchers to think harder has never changed behaviour; but the capacity it requires can be cultivated. In a study of undergraduates, critical thinking was found to mediate the relation between trust and reliance on generative AI, turning thoughtless acceptance into reflective, cautious, and selective use, so that trust did not issue in overreliance where critical engagement intervened (Hou et al. 2025). The disposition is teachable, and it has a known moderator: those more disposed to effortful thought benefit most from interventions that interrupt automatic acceptance (Buçinca et al. 2021). Yet one can contest an answer only in a field whose standards one has already internalised, and the novice cannot productively distrust what they are not yet equipped to judge. Calibrated reliance, therefore, presupposes some formation already in place; it protects the formed scholar more readily than it forms the unformed one. That is why the early stages of training are exactly where delegation should be most restrained, and why the burden for those not yet formed falls on the institutional level, with an embodied programme, rather than on the individual or design stance.

### 6.2. The Challenges of Embodied Scholarship Governance (*Outside the Loop*)

We can think about several institutional incentives for re-embodying scholarship. What is needed instead are arrangements that make the unautomatable activities of scholarship legible to the reward system, so that what a model cannot do becomes what the profession visibly prizes.[15] As we suggested in the Introduction, what is hardest to automate may gain in value, but only if institutions come to recognise it, and nothing guarantees that the recognition will be fairly distributed. Three lines follow, converging rather than competing. We offer them as directions entailed by the diagnosis, not as a programme we can warrant in detail; each raises a design problem we name where it arises, and the constraint that unifies them is to find proxies that the cheapening of text cannot capture.

The first is to reweight in-person scholarly participation in the institutions that govern careers. In fact, optimising the individual research process, even to the point of eliminating all cognitive friction, does not preserve what matters. What generates genuinely new ideas is often exposure to contexts outside one's immediate research environment: the peer met for the first time, the argument overheard between sessions, the discussion that refuses to take the form of a written summary. It is also, if not primarily, through live attendance that scholars come to know one another's reputation, in ways no publication record conveys. A reward system that counted conference participation explicitly in hiring and promotion, that asked authors to defend their submissions orally[16] before one or two invited readers prior to acceptance, and that restored the substantive viva – e.g., in the form of a simulated/mock lecture followed by structured questioning – in tenure and hiring procedures would recognise this dimension of scholarly life. None of these practices is new. Some were normal in the recent past and are still retained in parts of the academic world.

We should immediately complicate this, however. Simply giving more weight to conference participation in hiring and promotion is not, by itself, a way of rewarding the unautomatable. Conference programmes are themselves shaped by selection mechanisms that need not track the quality of in-person scholarly exchange at all. Acceptance is often decided on the basis of abstracts that an LLM can now write with reasonable fluency, programme committees are themselves susceptible to prestige cues, and the Matthew Effect (Merton 1968) ensures that visibility on programmes flows disproportionately to those already visible elsewhere. A metric that counts "conference talks given" would, predictably, be gamed in the same direction as publication counts. What needs rewarding is the quality of the talk and of the exchange surrounding it, and it is genuinely unclear how that quality should be measured at scale: certainly not by counting acceptances, and not obviously by any other proxy that lends itself to administrative use. We flag this as a substantive open question rather than waving it off as a solution.

A second set of interventions concerns what the system counts when it counts outputs. Not surprisingly, tying career advancement to publication volume creates (rational) incentives to maximise production at the expense of depth, and AI now eliminates the labour cost of volume across much of research. The remedy is not itself new: the principles articulated in the San Francisco Declaration on Research Assessment (DORA 2013) and the Leiden Manifesto (Hicks et al. 2015) were formulated for precisely this kind of pathology, though before AI made it acute. An update may be necessary. We want to be careful here, however, because the conclusion that academia should "rethink its metrics" is easier to state than to act on. The pull of metrification is not just that the current proxies are bad; it is that proxies of any kind, once they become the medium in which careers are evaluated, tend to crowd out the values they were introduced to track. Therefore, even if DORA and the Leiden Manifesto are right that h-index-style measures are crude, there is no guarantee that more sophisticated metrics will avoid the same fate. The risk is that each successive reform produces a new proxy that is initially better, then gamed, then capture-prone, and ultimately a fresh source of the very value displacement it was meant to remedy. We therefore treat the project of replacing current metrics as important but as harder than it tends to look.

A practice already in use points in a promising direction. Some funders and selection committees ask candidates to present a small number of publications they regard as their best, and assess them for their substance rather than counting them. The German Research Foundation (DFG) has limited the number of publications that may be listed in proposals and CVs since 2010, explicitly in the name of quality over quantity, and no longer asks for impact factors or h-indices (DFG 2022); narrative and selective CV formats are now widely discussed as tools for fairer assessment (Strinzel et al. 2021; CoARA 2022).

A third, final intervention follows from the same cost collapse, this time on the review side. Writing a decent paper, one that at least can be considered for publication, has become, in many areas, cognitively cheap. Journals are, therefore, absorbing a corresponding rise in submissions. Scholars who once received one review request per week now receive two or three, leading to more papers being reviewed with less attention per paper. The submission process should be made more expensive and the revenue used to pay reviewers for their work, an old and contested proposal, already the norm in parts of science, but under current conditions, increasingly unavoidable.[17]

We are aware of the distributional concerns that this last proposal (and potentially all the re-embodiment) raises. The existing Article Processing Charge (APC) model has already introduced significant inequities, disadvantaging early-career scholars and under-resourced institutions (Khoo 2019; Tennant et al. 2016). A submission-fee regime would compound that problem. A responsible version of this proposal would need waiver schemes tied to national income and institutional capacity. Without those safeguards, alternative instruments should be preferred, such as review credits that can be used to submit without fees, or publicly funded review infrastructure, analogous to how some European funders now support open-access publishing.

The submission process should be made more costly in non-monetary terms. Journals might, for instance, attach reputational or temporal consequences to clearly unsuitable submissions, such as temporary submission bans following (repeated) desk rejections. That's what happens when submitting an unsuccessful proposal for an ERC grant (up to a 2-year ban). Another possibility would be a refundable submission fee, returned to authors whose papers pass initial screening or are accepted, perhaps in the form of submission tokens. Such mechanisms would aim to discourage excessive or bad-faith submissions, including those from predatory or fabricated authors, while avoiding a system that simply prices out less-resourced scholars.

None of these institutional measures is sufficient alone, and each can be gamed, but all three serve a single end, which is to make the academic system reward the scholar who drives rather than the one who is driven. That is the institutional form of the stance described in §6.1, and it is the answer that this paper owes its title: second scholarship. If the hardest thing to automate tends, over time, to become the most visibly prized, then what is left for us is to become the kind of researcher capable of that unautomatable work, and capable of defending it: of standing, in person and in real time, behind a claim one has actually reasoned one's way to. This is not a nostalgic attachment to the unautomatable, but the one path that does not buy efficiency at the cost of degrading research, for the capacities that resist automation are the very ones whose exercise forms the researcher; to delegate them is, at once, to surrender what will be most valued and to erode the formation on which research depends.

After the default to trust is disrupted and the embodied conditions of scholarship are rebuilt, what is left for us is therefore not less than before but more demanding: to remain answerable for what we claim, and to grow into the higher standard of mastery that better instruments raise rather than retire. That answerability – the capacity to own and defend what only a formed judgement could have produced

– is what is left for us, and, in the end, what the profession will have most reason to prize.

## 7. Conclusion

The threat that GenAI poses to research is not bad output. It is good output obtained the wrong way. Because formation is constitutive of research and not merely instrumental to it, a discipline can improve its deliverables while degrading itself.

Against the reigning answer — keep a human in the loop — we have argued that supervision is a self-undermining paradigm. One cannot supervise one's way to a competence one never built, so a policy of oversight spends the capital of formation without replenishing it. We introduced the hermeneutic drift failure to name its endpoint: distanciation without appropriation, an understanding that is the model's account rather than an interpretation the reader could produce and defend.

In our opinion, the alternative is neither unplugging, since in an onlife world there is no offline to return to, nor surrender. It is 'second scholarship': the re-appropriation of the craft after critique, chosen in full knowledge of the shortcut it declines, grounded in the four sources and warrants that cannot be automated — tacit knowledge, personal commitment, socialisation, deep reading — and enacted as re-embodiment. After the default to trust is disrupted and the embodied conditions of scholarship are rebuilt, what is left for us is therefore a more demanding form of mastery. One which asks researchers to remain answerable for what they claim, to know when AI may assist and when it would displace formation, and to grow into the higher standard of judgement that better instruments raise rather than retire.

We finally suggest that it is precisely the capacity to own and defend what only a formed judgement could have produced what, in the end, the profession will have most reason to prize.

**Conflicts of Interest**

The authors declare no conflicts of interest.

**Data Availability Statement**

Data sharing is not applicable to this article as no datasets were generated or analysed during the current study.

## References

Agrawal, A. K., J. McHale, and A. Oettl. 2026. "AI in Science." Working Paper No. 34953. Working Paper Series. National Bureau of Economic Research, March. https://doi.org/10.3386/w34953.

Aristotle. 2009. *The Nicomachean Ethics*. Translated by D. Ross, revised by L. Brown. Oxford University Press.

Autor, D. 2014. "Polanyi's Paradox and the Shape of Employment Growth." Working Paper No. 20485. Working Paper Series. National Bureau of Economic Research, September. https://doi.org/10.3386/w20485.

Autor, D. H. 2015. "Why Are There Still So Many Jobs? The History and Future of Workplace Automation." *Journal of Economic Perspectives* 29 (3): 3–30. https://doi.org/10.1257/jep.29.3.3.

Bainbridge, L. 1983. "Ironies of Automation." In *Analysis, Design and Evaluation of Man–Machine Systems*, edited by G. Johannsen and J. E. Rijnsdorp. Pergamon. https://doi.org/10.1016/B978-0-08-029348-6.50026-9.

Bisbee, J., J. D. Clinton, C. Dorff, B. Kenkel, and J. M. Larson. 2024. "Synthetic Replacements for Human Survey Data? The Perils of Large Language Models." *Political Analysis* 32 (4): 401–416. https://doi.org/10.1017/pan.2024.5.

Bjork, E. L., and R. A. Bjork. 2011. "Making Things Hard on Yourself, but in a Good Way: Creating Desirable Difficulties to Enhance Learning." In *Psychology and the Real World: Essays Illustrating Fundamental Contributions to Society*. Worth Publishers.

Bjork, R. A., and E. L. Bjork. 2020. "Desirable Difficulties in Theory and Practice." *Journal of Applied Research in Memory and Cognition* (Netherlands) 9 (4): 475–479. https://doi.org/10.1016/j.jarmac.2020.09.003.

Boyer, E. L. 1990. *Scholarship Reconsidered: Priorities of the Professoriate*. Carnegie Foundation for the Advancement of Teaching.

Broek, P. van den. 2010. "Using Texts in Science Education: Cognitive Processes and Knowledge Representation." *Science* (New York, N.Y.) 328 (5977): 453–456. https://doi.org/10.1126/science.1182594.

Buçinca, Z., M. B. Malaya, and K. Z. Gajos. 2021. "To Trust or to Think: Cognitive Forcing Functions Can Reduce Overreliance on AI in AI-Assisted Decision-Making." *Proceedings of the ACM on Human-Computer Interaction* 5 (CSCW1): 1–21. https://doi.org/10.1145/3449287.

Buzási, Á. 2022. "Paul Ricœur and the Idea of Second Naivety: Origins, Analogues, Applications." *Études Ricoeuriennes / Ricoeur Studies* 13 (2): 39–58.

Carter, J. A. 2020. "Intellectual Autonomy, Epistemic Dependence and Cognitive Enhancement." *Synthese* 197 (7): 2937–2961. https://doi.org/10.1007/s11229-017-1549-y.

Casner, S. M., R. W. Geven, M. P. Recker, and J. W. Schooler. 2014. "The Retention of Manual Flying Skills in the Automated Cockpit." *Human Factors* 56 (8): 1506–1516. https://doi.org/10.1177/0018720814535628.

Cassinadri, G. 2024. "ChatGPT and the Technology-Education Tension: Applying Contextual Virtue Epistemology to a Cognitive Artifact." *Philosophy & Technology* 37 (1): 14. https://doi.org/10.1007/s13347-024-00701-7.

Chapuis, K., P. Taillandier, and A. Drogoul. 2022. "Generation of Synthetic Populations in Social Simulations: A Review of Methods and Practices." *Journal of Artificial Societies and Social Simulation* 25 (2): 6.

CoARA (Coalition for Advancing Research Assessment). 2022. *Agreement on Reforming Research Assessment.* https://coara.eu.

Collins, H. M. 1974. "The TEA Set: Tacit Knowledge and Scientific Networks." *Science Studies* 4 (2): 165–185. https://doi.org/10.1177/030631277400400203.

Collins, H. 1990. *Artificial Experts: Social Knowledge and Intelligent Machines*. The MIT Press. https://direct.mit.edu/books/monograph/4797/Artificial-ExpertsSocial-Knowledge-and-Intelligent.

Collins, H. 2013. *Tacit and Explicit Knowledge.* University of Chicago Press.

Collins, H. 2025. "Why Artificial Intelligence Needs Sociology of Knowledge: Parts I and II." *AI & SOCIETY* 40 (3): 1249–1263. https://doi.org/10.1007/s00146-024-01954-8.

Collins, H., and S. Thorne. 2026. "Large Language Models and Scientific Discourse: Where's the Intelligence?" *Synthese* 207 (4): 160. https://doi.org/10.1007/s11229-026-05531-y.

Csibra, G., and G. Gergely. 2009. "Natural Pedagogy." *Trends in Cognitive Sciences* 13 (4): 148–153. https://doi.org/10.1016/j.tics.2009.01.005.

Delgado, P., C. Vargas, R. Ackerman, and L. Salmerón. 2018. "Don't Throw Away Your Printed Books: A Meta-Analysis on the Effects of Reading Media on Reading Comprehension." *Educational Research Review* 25 (November): 23–38. https://doi.org/10.1016/j.edurev.2018.09.003.

DeStefano, D., and J.-A. LeFevre. 2007. "Cognitive Load in Hypertext Reading: A Review." *Computers in Human Behavior*, Including the Special Issue: Avoiding Simplicity, Confronting Complexity: Advances in Designing Powerful Electronic Learning Environments, vol. 23 (3): 1616–1641. https://doi.org/10.1016/j.chb.2005.08.012.

Dewey, J. 1910. *How We Think*. D. C. Heath. https://doi.org/10.1037/10903-000.

DORA. 2013. *San Francisco Declaration on Research Assessment.* https://sfdora.org/read/.

Douglas, H. 2024. "On the Necessity of Embodiment for Reasoning." *The American Philosophical Association* 23 (2).

Dreyfus, H., S. E. Dreyfus, and T. Athanasiou. 1986. *Mind Over Machine*. Simon and Schuster.

Fan, Y., L. Tang, H. Le, et al. 2025. "Beware of Metacognitive Laziness: Effects of Generative Artificial Intelligence on Learning Motivation, Processes, and Performance." *British Journal of Educational Technology* 56 (2): 489–530. https://doi.org/10.1111/bjet.13544.

Felgenbaum, E. A. 1977. "The Art of Artificial Intelligence: Themes and Case Studies of Knowledge Engineering." *Proceedings of the 5th International Joint Conference on Artificial Intelligence - Volume 2* (San Francisco, CA, USA), IJCAI'77, August 22, 1014–1029.

Felski, R. 2015. *The Limits of Critique*. University of Chicago Press.

Floridi, L. 2014. *The Fourth Revolution: How the Infosphere Is Reshaping Human Reality*. Oxford University Press UK.

Floridi, L. 2015. *The Onlife Manifesto - Being Human in a Hyperconnected Era*. Springer. https://doi.org/10.1007/978-3-319-04093-6.

Fonagy, P., and E. Allison. 2014. "The Role of Mentalizing and Epistemic Trust in the Therapeutic Relationship." *Psychotherapy* (US) 51 (3): 372–380. https://doi.org/10.1037/a0036505.

Gerlich, M. 2025. "AI Tools in Society: Impacts on Cognitive Offloading and the Future of Critical Thinking." *Societies* 15 (1). https://doi.org/10.3390/soc15010006.

Ginzburg, C. 2023. *Miti emblemi spie. Morfologia e storia. Nuova ediz*. Adelphi.

Goldberg, S. 2011. "The Division of Epistemic Labor." *Episteme* 8 (1): 112–125. https://doi.org/10.3366/epi.2011.0010.

Goldberg, S. C. 2020. "Trust and Reliance." In *The Routledge Handbook of Trust and Philosophy*. Routledge.

Grossmann, M. 2021. *How Social Science Got Better: Overcoming Bias with More Evidence, Diversity, and Self-Reflection*. Oxford University Press.

Guillory, J. 2010. "Close Reading: Prologue and Epilogue." *ADE Bulletin* 149: 8–14.

Hamming, R. W. 1986. "You and Your Research." Transcript of a talk given at the Bell Communications Research Colloquium Seminar, 7 March 1986. https://fs.blog/great-talks/richard-hamming-your-research/.

Hardwig, J. 1991. "The Role of Trust in Knowledge." *The Journal of Philosophy* 88 (12): 693–708. https://doi.org/10.2307/2027007.

Heersmink, R., B. de Rooij, M. J. Clavel Vázquez, and M. Colombo. 2024. "A Phenomenology and Epistemology of Large Language Models: Transparency, Trust, and Trustworthiness." *Ethics and Information Technology* 26 (3): 41. https://doi.org/10.1007/s10676-024-09777-3.

Hicks, D., P. Wouters, L. Waltman, S. de Rijcke, and I. Rafols. 2015. "Bibliometrics: The Leiden Manifesto for Research Metrics." *Nature* 520: 429–431.

Hou, C., G. Zhu, V. Sudarshan, F. S. Lim, and Y. S. Ong. 2025. "Measuring Undergraduate Students' Reliance on Generative AI during Problem-Solving: Scale Development and Validation." *Computers & Education* 234 (September): 105329. https://doi.org/10.1016/j.compedu.2025.105329.

Howells, J. 1996. "Tacit Knowledge." *Technology Analysis & Strategic Management* 8 (2): 91–106. https://doi.org/10.1080/09537329608524237.

Kambhampati, S. 2021. "Polanyi's Revenge and AI's New Romance with Tacit Knowledge." *Communications of the ACM* 64 (2): 31–32. https://doi.org/10.1145/3446369.

Keller, T. A., R. A. Mason, A. E. Legg, and M. A. Just. 2024. "The Neural and Cognitive Basis of Expository Text Comprehension." *Npj Science of Learning* 9 (1): 21. https://doi.org/10.1038/s41539-024-00232-y.

Khoo, S. Y.-S. 2019. "Article Processing Charge Hyperinflation and Price Insensitivity: An Open Access Sequel to the Serials Crisis." *LIBER Quarterly: The Journal of the Association of European Research Libraries* 29 (1): 1–18. https://doi.org/10.18352/lq.10280.

Koskinen, I. 2024. "We Have No Satisfactory Social Epistemology of AI-Based Science." *Social Epistemology* 38 (4): 458–475. https://doi.org/10.1080/02691728.2023.2286253.

Kosmyna, N., E. Hauptmann, Y. T. Yuan, et al. 2025. "Your Brain on ChatGPT: Accumulation of Cognitive Debt When Using an AI Assistant for Essay Writing Task." arXiv:2506.08872. Preprint, arXiv, December 31. https://doi.org/10.48550/arXiv.2506.08872.

Krenn, M., R. Pollice, S. Y. Guo, et al. 2022. "On Scientific Understanding with Artificial Intelligence." *Nature Reviews. Physics* 4 (12): 761–769. https://doi.org/10.1038/s42254-022-00518-3.

Kuhn, T. S. 1962. *The Structure of Scientific Revolutions.* University of Chicago Press.

Kusumegi, K., X. Yang, P. Ginsparg, M. de Vaan, T. Stuart, and Y. Yin. 2025. "Scientific Production in the Era of Large Language Models." *Science* 390 (6779): 1240–1243. https://doi.org/10.1126/science.adw3000.

Lakatos, I. 1970. "Falsification and the Methodology of Scientific Research Programmes." In *Criticism and the Growth of Knowledge: Proceedings of the International Colloquium in the Philosophy of Science, London, 1965*, edited by Alan Musgrave and Imre Lakatos, vol. 4. Cambridge University Press. https://doi.org/10.1017/CBO9781139171434.009.

Lave, J., and E. Wenger. 1991. *Situated Learning: Legitimate Peripheral Participation*. Cambridge University Press.

Lazaros, K., A. G. Vrahatis, and S. Kotsiantis. 2026. "Human-in-the-Loop Artificial Intelligence: A Systematic Review of Concepts, Methods, and Applications." *Entropy* 28 (4): 377. https://doi.org/10.3390/e28040377.

Lee, H.-P., A. Sarkar, L. Tankelevitch, et al. 2025. "The Impact of Generative AI on Critical Thinking: Self-Reported Reductions in Cognitive Effort and Confidence Effects From a Survey of Knowledge Workers." *Proceedings of the 2025 CHI Conference on Human Factors in Computing Systems* (New York, NY, USA), CHI '25, April 25, 1–22. https://doi.org/10.1145/3706598.3713778.

Lee, J. D., and K. A. See. 2004. "Trust in Automation: Designing for Appropriate Reliance." *Human Factors* 46 (1): 50–80. https://doi.org/10.1518/hfes.46.1.50_30392.

Liang, W., Y. Zhang, Z. Wu, et al. 2025. "Quantifying Large Language Model Usage in Scientific Papers." *Nature Human Behaviour* 9 (12): 2599–2609. https://doi.org/10.1038/s41562-025-02273-8.

Liu, J., J. Pei, J. Huang, et al. 2026. "The Last Human-Written Paper: Agent-Native Research Artifacts." arXiv:2604.24658. Preprint, arXiv, May 19. https://doi.org/10.48550/arXiv.2604.24658.

Longino, H. E. 1990. *Science as Social Knowledge : Values and Objectivity in Scientific Inquiry*. Princeton University Press.

Loru, E., J. Nudo, N. Di Marco, et al. 2025. "The Simulation of Judgment in LLMs." *Proceedings of the National Academy of Sciences* 122 (42): e2518443122. https://doi.org/10.1073/pnas.2518443122.

Mann, S. P., M. Aboy, J. J. Seah, et al. 2025. "AI and the Future of Academic Peer Review." arXiv.Org, September 17. https://arxiv.org/abs/2509.14189v3.

McIntyre, L. 2019. *The Scientific Attitude: Defending Science from Denial, Fraud, and Pseudoscience*. MIT Press.

Mercier, H., and D. Sperber. 2017. *The Enigma of Reason*. Harvard University Press.

Merton, R. K. 1968. "The Matthew Effect in Science." *Science* 159: 56–63.

Naddaf, M. 2025. "AI Linked to Explosion of Low-Quality Biomedical Research Papers." *Nature* 641 (8065): 1080–1081. https://doi.org/10.1038/d41586-025-01592-0.

Naddaf, M. 2026. "Why Paying Peer Reviewers Works, According to a Journal's Editor-in-Chief." *Nature*, ahead of print, July 1. https://doi.org/10.1038/d41586-026-01973-z.

Negel, J. 2013. "Zweite Naivität: Begriffsgeschichtliche und systematische Erwägungen zu einem vielbemühten, aber selten verstandenen Konzept." In *Welt als Gabe: Hermeneutische Grenzgänge zwischen Theologie und Phänomenologie*, 259–288. Aschendorff Verlag.

Nguyen, C. T. 2022. "Trust as an Unquestioning Attitude." *Oxford Studies in Epistemology* 7: 214–244.

Novelli, C., J. Argota Sánchez-Vaquerizo, D. Helbing, A. Rotolo, and L. Floridi. 2026. "A Replica for Our Democracies? On Using Digital Twins to Enhance Deliberative Democracy." *AI & SOCIETY* 41 (3): 1783–1801. https://doi.org/10.1007/s00146-025-02511-7.

Novelli, C., J. Argota Sanchez-Vaquerizo, J. Cyr, et al. 2026. "Fake Plastic Voters: When Political Parties Can Use AI-Simulated Focus Groups." arXiv:2606.00043. Preprint, arXiv, April 28. https://doi.org/10.48550/arXiv.2606.00043.

Nudo, J., M. E. Pandolfo, E. Loru, M. Samory, M. Cinelli, and W. Quattrociocchi. 2026. "Generative Exaggeration in LLM Social Agents: Consistency, Bias, and Toxicity." *Online Social Networks and Media* 51 (January): 100344. https://doi.org/10.1016/j.osnem.2025.100344.

Ouyang, L., J. Wu, X. Jiang, et al. 2022. "Training Language Models to Follow Instructions with Human Feedback." *Advances in Neural Information Processing Systems* 35: 27730–27744.

Page, S., N. Cantor, and E. Lewis. 2019. *The Diversity Bonus : How Great Teams Pay Off in the Knowledge Economy*. Princeton University Press.

Peirce, C. S. 1979. *Collected Papers of Charles Sanders Peirce, Vol. 7: Science and Philosophy / Vol. 8: Reviews, Correspondence and Bibliography*. Edited by Arthur W. Burks. Harvard University Press.

Peters, U., and B. Chin-Yee. 2025. "Generalization Bias in Large Language Model Summarization of Scientific Research." *Royal Society Open Science* 12: 241776. https://doi.org/10.1098/rsos.241776.

Polanyi, M. 1958. *Personal Knowledge: Towards a Post-Critical Philosophy*. Routledge.

Polanyi, M. 1968. "Logic and Psychology." *American Psychologist* (US) 23 (1): 27–43. https://doi.org/10.1037/h0037692.

Polanyi, M. 2009. *The Tacit Dimension*. Edited by Amartya Sen. University of Chicago Press. https://press.uchicago.edu/ucp/books/book/chicago/T/bo6035368.html.

Resnik, D. B., and M. Hosseini. 2025. "The Ethics of Using Artificial Intelligence in Scientific Research: New Guidance Needed for a New Tool." *AI and Ethics* 5 (2): 1499–1521. https://doi.org/10.1007/s43681-024-00493-8.

Richards, I. A. 1929. *Practical Criticism: A Study of Literary Judgment*. Kegan Paul.

Ricoeur, P. 1966. *The Symbolism of Evil*. Harper & Row.

Rolin, K. 2020. "Trust in Science." In *The Routledge Handbook of Trust and Philosophy*. Routledge.

Sharma, M., M. Tong, T. Korbak, et al. 2025. "Towards Understanding Sycophancy in Language Models." arXiv:2310.13548. Preprint, arXiv, May 10. https://doi.org/10.48550/arXiv.2310.13548.

Sher, G. 2010. "Epistemic Friction: Reflections on Knowledge, Truth, and Logic." *Erkenntnis* 72 (2): 151–176. https://doi.org/10.1007/s10670-009-9202-x.

Sorkin, D. 1983. "Wilhelm von Humboldt: The Theory and Practice of Self-Formation (Bildung), 1791–1810." *Journal of the History of Ideas* 44 (1): 55–73.

Stadler, M., M. Bannert, and M. Sailer. 2024. "Cognitive Ease at a Cost: LLMs Reduce Mental Effort but Compromise Depth in Student Scientific Inquiry." *Computers in Human Behavior* 160 (November): 108386. https://doi.org/10.1016/j.chb.2024.108386.

Strinzel, M., J. Brown, W. Kaltenbrunner, S. de Rijcke, and M. Hill. 2021. "Ten Ways to Improve Academic CVs for Fairer Research Assessment." *Humanities and Social Sciences Communications* 8: 251. https://doi.org/10.1057/s41599-021-00929-0.

Talwar, V., J. Lavoie, and A. M. Crossman. 2022. "Socialization of Lying Scale: Development and Validation of a Parent Measure of Socialization of Truth and Lie-Telling Behavior." *Applied Developmental Science* 26 (3): 553–566. https://doi.org/10.1080/10888691.2021.1927732.

Tao, T. 2026. "Kepler, Newton, and the True Nature of Mathematical Discovery." Interview by D. Patel. *The Dwarkesh Podcast*, March 2026. https://www.dwarkesh.com/p/terence-tao.

Tennant, J. P., F. Waldner, D. C. Jacques, P. Masuzzo, L. B. Collister, and C. H. J. Hartgerink. 2016. "The Academic, Economic and Societal Impacts of Open Access: An Evidence-Based Review." *F1000Research* 5 (September): 632. https://doi.org/10.12688/f1000research.8460.3.

Thornton, T. 2006. "Tacit Knowledge as the Unifying Factor in Evidence Based Medicine and Clinical Judgement." *Philosophy, Ethics, and Humanities in Medicine* 1 (March): 2. https://doi.org/10.1186/1747-5341-1-2.

Tomasello, M. 2018. "How Children Come to Understand False Beliefs: A Shared Intentionality Account." *Proceedings of the National Academy of Sciences* 115 (34): 8491–8498. https://doi.org/10.1073/pnas.1804761115.

Turner, S. 2022. "Polanyi and Tacit Knowledge." In *The Routledge Handbook of Philosophy and Implicit Cognition*. Routledge.

Vaswani, A., N. Shazeer, N. Parmar, et al. 2017. "Attention Is All You Need." *Advances in Neural Information Processing Systems* 30: 5998–6008.

Vallor, S. 2016. *Technology and the Virtues: A Philosophical Guide to a Future Worth Wanting*. Oxford University Press.

Westwood, S. J. 2025. "The Potential Existential Threat of Large Language Models to Online Survey Research." *Proceedings of the National Academy of Sciences* 122 (47): e2518075122. https://doi.org/10.1073/pnas.2518075122.

Wiley. 2025. *ExplanAItions 2025: The Evolution of AI in Research*. John Wiley & Sons. https://www.wiley.com/en-us/about-us/ai-resources/ai-study/key-findings/.

Wolf, M., and M. Barzillai. 2009. "The Importance of Deep Reading." *Educational Leadership* 66 (6): 32–37.

Wolf, M. 2010. *Proust and the Squid: The Story and Science of the Reading Brain*. Harper Perennial.

Wolf, M. 2018. *Reader, Come Home: The Reading Brain in a Digital World*. Harper, an imprint of HarperCollinsPublishers.

Yang, Z., Z. Zhang, Z. Zheng, et al. 2025. "OASIS: Open Agent Social Interaction Simulations with One Million Agents." arXiv:2411.11581. Preprint, arXiv, March 23. https://doi.org/10.48550/arXiv.2411.11581.

Zhu, L., Y. Lai, J. Xie, et al. 2025. "Evaluating the Potential Risks of Employing Large Language Models in Peer Review." *Clinical and Translational Discovery* 5 (4): e70067. https://doi.org/10.1002/ctd2.70067.

---

[1] Some scientists argue, optimistically, that AI is comparable to a computational microscope that reveals structures or dynamics inaccessible to us, or as a resource of inspiration for genuinely new ideas (Krenn et al. 2022, 4).

[2] Responsibility and answerability occupy distinct registers. Responsibility concerns an agent's fitness for praise or blame and her standing within a community of obligation. Answerability denotes a belief or claim's responsiveness to evidence and reasons, admitting rational correction rather than moral sanction.

[3] To re-embody research is not to step away from digital technology. That prescription will not happen. In a world that is by now thoroughly onlife, in which the boundary between online and offline has dissolved to the point where it is no longer sensible to ask on which side of it one stand (Floridi 2014; Floridi 2015), there is no offline to which one could retreat. Re-embodiment, as we use it, is the continued exercise of the formative activities that AI delegation displaces, carried on within the onlife condition rather than against it.

[4] By 'scholarship' we mean the full range of research activities, not the humanities alone.

[5] The term adapts Ricoeur's *second naïveté* (Ricoeur 1966, 349–352). Its three-stage movement – first naïveté, critical distanciation, and a mediated return – is not confined to religious symbolism: Buzási (2022) argues that it extends to interpretation in general, and it is that generalised form we carry to

scholarly practice. The phrase has a longer history: Peter Wust coined it in 1925 and Ernst Simon used it independently in 1931, and Ricoeur most likely took it from Gabriel Marcel; on this genealogy, and on the German *zweite Naivität* see Negel (2013). Second scholarship is neither the postcritique of literary studies (Felski 2015), whose critical moment is the hermeneutics of suspicion rather than automation, nor a member of Boyer's (1990) typology of the scholarships of discovery, integration, application, and teaching, which sorts kinds of academic work rather than a stage in the formation of judgement.

[6] There are empirical studies that point at the limit that such an environment may have (Nudo et al. 2026), but it is not our purpose here to defend it. We have been very skeptical about them as well (Novelli, Argota Sánchez-Vaquerizo, et al. 2026; Novelli, Sanchez-Vaquerizo, et al. 2026).

[7] This is taken from Charles Peirce's account of how scientific questions arise (Peirce 1979).

[8] Cognitive friction operates alongside other forms of friction in scholarly practice – e.g., instrumental, dialectical, communal – and across the disciplines these forms are intertwined rather than insulated. On the broader notion of epistemic friction, including the multiple sources from which it arises, see Sher (2010).

[9] The point recalls Ginzburg's account, in his discussion of the "paradigma indiziario" (clue-based mode of knowledge), that is, of mute forms of knowledge that are functional to evidence but whose rules cannot be fully formalized or stated (Ginzburg 2023).

[10] We thank Professor Heather Douglas for drawing our attention to this point. While we may differ on whether this kind of epistemic disturbance is best understood as tacit knowledge, we share her view that it is epistemically significant and that its role in inquiry depends on embodied reasoning. See also Douglas (2024).

[12] At the same time, peer-reviewed work cited in §3, on metacognitive laziness and on the erosion of critical thinking under high confidence in the system (Fan et al. 2025; Lee et al. 2025), already points in the same direction with firmer footing.

[13] And even more, over time, what begins as a contextual value can harden into an apparently neutral epistemic norm. So, for instance, the early modern interest in controlling and manipulating nature: this practical, contextually motivated goal promoted mechanistic models of the world, which gradually became the taken-for-granted framework within which good explanation was defined (Longino 1990, 94 ff.).

[14] We use deep reading in the specific sense of the cognitive neuroscience of reading (Wolf 2018), the capacity supported by the reading circuit, rather than as a synonym for the humanities' own term of art, close reading, which denotes disciplined attention to a text's language and structure (Richards 1929; Guillory 2010). The two are complementary: close reading is among the practices that exercise the deep reading circuit, and the circuit is the capacity that close reading presupposes. We foreground the capacity because our argument turns on its erosion, a claim about formation that the method itself does not make. So understood, close reading is a paradigm of the first-personal engagement that second scholarship re-appropriates.

[15] Re-embodiment, as defined in §4, is primarily a matter of doing one's own formative work, not of physical location. The in-person arrangements proposed below matter because, as the socialisation

argument of §4.3 showed, co-presence remains the setting in which trust, reputation, and collective tacit knowledge are most reliably formed and displayed.

[16] The same logic applies to teaching: in many disciplines, university examination should return to, or retain, the oral form, in which understanding must be articulated and defended in real time rather than produced through an interface.

[17] A recent experiment by Naddaf (2026) found that paying reviewers improved the efficiency of the process while maintaining review quality .